\documentclass[sigconf, screen, review]{acmart}

\usepackage{subcaption}

\AtBeginDocument{%
  \providecommand\BibTeX{{%
    \normalfont B\kern-0.5em{\scshape i\kern-0.25em b}\kern-0.8em\TeX}}}

\setcopyright{acmcopyright}
\copyrightyear{2021}
\acmYear{2021}
\acmDOI{10.1145/1122445.1122456}

\acmConference[FDG '21]{FDG '21: Foundations of Digital Games}{August 03--06, 2021}{Montreal, Canada}
\acmBooktitle{FDG '21: Foundations of Digital Games,
  August 03--06, 2021, Montreal, Canada}
\acmPrice{15.00}
\acmISBN{978-1-4503-XXXX-X/18/06}

\begin{document}

\title{Game Mechanic Alignment Theory}%% and Discovery}

% \author{Anonymous}
% \email{anonymous@email.com}
% \affiliation{%
%   \institution{Anonymous Institute}
%   \city{Anonymous City}
%   \state{Anonymous State}
%   \country{Anonymous Country}
% }
\author{Michael Cerny Green}
\email{mike.green@nyu.edu}
\affiliation{%
  \institution{New York University | OriGen.AI}
  \city{New York City}
  \state{New York}
  \country{USA}
}

\author{Ahmed Khalifa}
\email{ahmed@akhalifa.com}
\affiliation{%
  \institution{Game Innovation Lab | Modl.ai}
  \city{New York City}
  \state{New York}
  \country{USA}
}

\author{Philip Bontrager}
\email{pbontrager@gmail.com}
\affiliation{%
  \institution{New York University}
  \city{New York City}
  \state{New York}
  \country{USA}
}

\author{Rodrigo Canaan}
\email{rmc602@nyu.edu}
\affiliation{%
  \institution{New York University}
  \city{New York City}
  \state{New York}
  \country{USA}
}

\author{Julian Togelius}
\email{julian@togelius.com}
\affiliation{%
  \institution{New York University}
  \city{New York City}
  \state{New York}
  \country{USA}
}

\renewcommand{\shortauthors}{Green, et al.}

\begin{abstract}
  We present a new concept called Game Mechanic Alignment theory as a way to organize game mechanics through the lens of systemic rewards and agential motivations. By disentangling player and systemic influences, mechanics may be better identified for use in an automated tutorial generation system, which could tailor tutorials for a particular playstyle or player.
  Within, we apply this theory to several well-known games to demonstrate how designers can benefit from it, we describe a methodology for how to estimate ``mechanic alignment'', and we apply this methodology on multiple games in the GVGAI framework. 
  We discuss how effectively this estimation captures agential motivations and systemic rewards and how our theory could be used as an alternative way to find mechanics for tutorial generation.
\end{abstract}




\maketitle

\section{Introduction}

A player's first experience with a video game is often with its tutorial. Tutorials provide a way for a game designer to communicate with the player, to train them, and to help them understand the game's rules. Without this guidance, the player may become frustrated, unable to figure out how to play or feel a sense of progression. At worst, a tutorial is unhelpful and confusing. But if done correctly, a tutorial excites and encourages the player to keep playing.

To be effective teachers, tutorials need to contain the crucial bits of information needed to play, such as what the controls are, how to win, and how to lose. This information is typically defined by the game's mechanics, i.e. the events within the game triggered by game elements that impact the game state~\cite{sicart2008defining}. 
% If tutorials could be seen as a dialogue between player and designer, mechanics could be seen as communication between the player and the environment.
%julian does the information "come in" that form, or is it more that it can be described that way? or that "game mechanics" are the way we partition the information into parts, or "bits"? We should not give the impression that game mechanics exist because of tutorials
% mike done
``Critical mechanics'' are the mechanics that need to be triggered in order to win a level~\cite{green2020automatic}. Therefore, it makes logical sense that critical mechanics be explained within a tutorial. Previous work has 
%julian Please don't use the word "researches". Research is not a countable noun. 
%mike dont know how that snuck in there...
proposed several solutions to automatically find critical mechanics (coined ``critical mechanic discovery methods'') using uninformed~\cite{green2017press,green2018atdelfi} and informed~\cite{green2020automatic} tree search methods and a graph of game mechanic relationships.

However, player enjoyment cannot and should not be limited to a binary choice of ``winning'' and ``losing''. Players engage with games with their own biases and motivations, which impact the enjoyment they receive from play. For example, in the game Minecraft (Mojang 2007), a voxel-based open-world sandbox game, the designers may have intended for players to travel to the ``End,'' a
%julian Did the designers "intend" this? More like they gave the player the option to travel to the end.
%mike I added an extreme case about TNT (will add youtube video explaining this as a footnote). I used the phrase "may have intended" to show uncertainty at the designer's true intentions
dangerous zone of monsters and treacherous terrain, to defeat the Ender Dragon, which could act as the game's final boss. However, many players may choose to never travel to the End, instead selecting to build large castles and  design elegant structures that are aesthetically pleasing. In an extreme case, some players employ TNT in an explosive attempt to blow up as much of the rendered world as possible without crashing their server~\footnote{Example video of TNT explosion gameplay: https://www.youtube.com/watch?v=o9V2FOyD-GM}.
Admittedly, defeating the Ender Dragon does not explicitly end the game either, as the player can keep on playing in their world with the ability to slay the Ender Dragon repeatedly. 
% The challenge of the Ender Dragon can be entirely replaced with a form of virtual legos. 

Elias, Gutschera and Garfield~\cite{elias2012characteristics} distinguish between ``systemic'' and ``agential'' properties of game characteristics of a game. Systemic properties derive from the game's rules, and agential properties depend on the players. In this work, we consider that the choice of a player to trigger a mechanic can be made through a combination of systemic and agential factors: a mechanic provides ``systemic rewards'' if it results in the player obtaining an external (to the player) reward signal created by the designer to encourage certain behavior (e.g. points) or if it directly contributes to progression or winning. However, mechanics can also be pursued due to ``agential motivations'' that are internal to the player. These motivations may be aligned with or run against systemic rewards

% The game designer creates external, ``systemic'' rewards, created by the design to guide the player into behaving a certain way. However, a player has their own internal, ``agential'' motivations, which may be aligned with or run against these systemic rewards. 

By categorizing mechanics within a framework that provides spaces for both systemic and agential influence, we can better understand game mechanics and how to teach them to players. Automated tutorial generation~\cite{green2018atdelfi} is a relatively unexplored artificial intelligence application that could greatly benefit from such a framework. Critical mechanic discovery~\cite{green2020automatic}, i.e. the process to automatically find which mechanics to teach inside a tutorial, provides a useful family of methods to feed an automated tutorial generator. However, previous tutorial generation methods are limited and have drawbacks, chief among those being a reliance a complex game graph of mechanical relationships similar to the graphs created by Machinations framework\footnote{https://machinations.io/}~\cite{dormans2011simulating}.  These game graphs require innate knowledge of the game-in-question and how different mechanics relate to one another. Such analysis can be difficult without detailed insight about the game elements themselves, making them difficult for developers to quickly deploy them. These methods also do not leave room for differences between players, who may play the same level differently from one another depending on their goals.

This paper presents a \emph{Game Mechanic Alignment} theory, a framework in which mechanics can be analyzed in terms of agential motivations and systemic rewards.
This work includes examples of Game Mechanic Alignment applied to some well-known games, a methodology that can estimate mechanic alignment given playtrace data, and an application of this methodology on several video games in the General Video Game Artificial Intelligence (GVGAI) framework~\cite{perez2019general}. We conclude how the theory can be applied as input to automated tutorial generation systems as an alternative method to critical mechanic discovery.

\section{Game Mechanics and Tutorials}

Sicart defines a ``game mechanic'' as an event within the game that is fired by a game element that impacts the game's state~\cite{sicart2008defining}. Mechanics allow players to interact with and impact the game's state. Tutorials are meant to help players understand game mechanics and, ultimately, learn how to play with them. 

% A ``game mechanic'' can be defined as an event within the game that is fired by a game element that impacts the game's state~\cite{sicart2008defining}. Mechanics allow players to interact with and impact the game's state. Tutorials are meant to help players understand game mechanics and, ultimately, learn how to play with them. 

\subsection{Game Mechanics}
In addition to his definition of game mechanics, Sicart defines  ``core,'' ``primary,'' and ``secondary'' mechanics, which are also relevant to this work:
\begin{itemize}
\item Core mechanics are (repetitively) used by agents to achieve a systemically rewarded end-state.
\item Primary mechanics are the subset of core mechanics that can be directly applied by agents to solve challenges that lead to a desired end-state.
\item Secondary mechanics are the subset of core mechanics that make it easier for the agent to reach the desired end-state but are not essential like primary mechanics.
%julian I would really need an example here to grasp the difference
\end{itemize}
For example, in Super Mario Bros (Nintendo, 1985) World 1-1, Jumping and Running are both core mechanics. Jumping is a primary mechanic as you can't finish the level without it. Running is a secondary mechanic as it makes jumping easier and finishing the level faster, but it is not necessary to reach to the desired end state. Another example is Pacman (Namco 1980), where eating pellets is primary and eating ghosts is secondary.

Indeed core mechanics have been defined by others~\cite{salen2004rules,jarvinen2008games,adams2007game}, but lack a systemic perspective. For example, Salen and Zimmerman define them as ``the essential play activity players perform again and again in a game (...) however, in many games, the core mechanic is a compound activity composed of a suite of actions''~\cite{salen2004rules}. Jarvinen indirectly touches upon reward systems with ``the possible or preferred or encouraged means with which the player can interact with game elements as she is trying to influence the game state at hand towards the attainment of a goal''~\cite{jarvinen2008games}, but still does not tell us \emph{whose goals} and if the core mechanics are actually essential to attaining it.
%julian in the same way??
Although Sicart's definitions provide precision with this useful formalism, they admittedly lack the ability to classify game mechanics outside of those oriented around environmentally defined goal/reward structures. Our proposed framework attempts to cover some of this undefined space, providing a method to analyze mechanics through the lens of player and environment.

\subsection{Tutorials and Tutorial Generation}
Developers have experimented with multiple tutorial formats ~\cite{therrien2011get}. For simple games meant to be picked up and played quickly, mechanics tend to be intuitive: ``Press space to shoot'', ``Press up to jump'', and so on. As a result, these games usually lacked a formal tutorial. As game complexity increased and home consoles started to explode in popularity, formal tutorials became more common. Tutorials have since evolved to incorporate different design and presentation styles depending on the taste and conviction of game designers and their perception of players~\cite{green2017press}. Tutorials can adapt to different learning capabilities of the users who use them. Sheri Graner Ray~\cite{ray2010learning} discusses different \textit{knowledge acquisition styles}. Explorative Acquisition follows a ``learning by doing'' philosophy, while Modeling Acquisition, is about ``reading before doing.'' Green et al.~\cite{green2017press} proposes three different presentation styles within the same vein: \emph{Text}, \emph{Demonstrations}, and \emph{Well-Designed Experiences}.

Several projects have addressed challenges in automated tutorial generation, such as heuristic generation for Blackjack and Poker \cite{de2016generating,de2018flop,de2018texas} or quest/achievement generation in \emph{Minecraft}~\cite{alexander2017deriving}. Mechanic Miner~\cite{cook2013mechanic} evolves mechanics for 2D puzzle-platform games, using \emph{Reflection}\footnote{https://code.google.com/archive/p/reflections/} to find a new game mechanic then generate levels that utilize it.
%julian Is that source code link really for reflection as Cook uses it? I thought that was the Java reflection API?
%ahmed that is the definition of the word "reflection" for people who don't understand it.
The \emph{Gemini} system~\cite{summerville2017mechanics} takes game mechanics as input and performs static reasoning to find higher-level meanings about the game. Mappy~\cite{osborn2017automatic} can transform a series of button presses into a graph of room associations, transforming movement mechanics into level information for any Nintendo Entertainment System game. 

% \begin{figure}
%     \centering
%     \includegraphics[width=0.6\columnwidth]{images/etc/tutorial-card-zelda.png}
%     \caption{An AtDelfi generated tutorial for GVGAI's Zelda}
%     \label{fig:atdelfi}
% \end{figure}

The \emph{AtDelfi} system~\cite{green2018atdelfi} attempts to solve the challenge of automatically generating tutorials for video games using two different formats: text-based instructions and curated GIF demonstrations. This has been later expanded upon to include small sub-levels~\cite{green2018generating,khalifa2019intentional} and entire levels in Mario~\cite{green2020mario} and 2d arcade games~\cite{charity2020mech}. To develop these tutorials, each system requires an input set of game mechanics, referred to as the \emph{critical mechanics}: the set of mechanics that are necessary to trigger in order to win the level.
% Being able to automatically generate tutorials would be of large benefit to developers, as most tutorials are created by hand. 
% Automated tutorial generation could also augment fully automatic game generation.
% which so far has proven difficult as previous attempts so far have demonstrated that evaluating generated games for humans, without using human-like playing ability~\cite{nielsen2015towards,cook2014ludus} is not trivial.
%julian And uses these mechanics to create complete tutorials? Maybe say why this is necessary, as this is a key part of the story we're telling here?
% Figure~\ref{fig:atdelfi} displays a tutorial card generated by the system. 
In addition to presenting a method to automatically find critical mechanics, \emph{AtDelfi} also includes mechanics that give the player points or cause a loss.  \emph{Talin}~\cite{aytemiz2018talin} is a Unity-based tutorial generation system which dynamically presents information to a player based on their skill-level. Novice players will be presented with more information, whereas experienced players will be spared unneeded tooltips. Talin differentiates itself from AtDelfi in that it does not attempt to automatically discover which mechanics are critical, but instead which of the manually-selected mechanics need to be displayed for the user's consumption. 

\section{Player Behavior and Player Modeling}
By choosing to expose or omit certain game mechanics, a tutorial may cater to some player's motivations over others. In this work, we suggest that automated tutorial generative systems should incorporate not only different presentation formats and learning styles, but also playstyles .

\subsection{Player Behavior}
Ribbens et al. wrote that studying player behavior should not occur in isolation of the game environment~\cite{ribbens2009researching}. A player cannot play without an environment to interact within, and gameplay is intimately intertwined with the particular player. 

Gameplay environments are created and shaped by the game designer. Conscious design decisions can influence player behavior and therefore the player experience~\cite{bergstrom2013constructing}. But player behavior is not just influenced by the environment. Players carry their own biases into the game that influence their in-game behavior, stemming from individual motivations~\cite{kuhlman1975individual},  culture~\cite{bialas2014cultural}, and even age~\cite{tekofsky2013age}. By analyzing behaviors, one can categorize players into a taxonomy of playstyles~\cite{bartle2004designing,yee2002facets}, each category not being mutually exclusive of the rest.
Artificial intelligence systems may assist with player behavioral analysis~
\cite{horn2017ai}. Our work presents  a novel method to analyze player behavior by focusing on game mechanic usage during play.

\subsection{Player Modeling}
Player modeling is the study of computational models of players in games, including their incentives and behavior~\cite{yannakakis2013playermodeling}. It is often used to study and even mimic the styles of players. This can be done using methods such as supervised learning using real playtraces~\cite{ortega2013imitating,togelius2007towards} or utility-function formulation~\cite{holmgard2014evolvingpersonas,holmgaard2014personas,holmgard2015mcts,holmgaard2018automated}. Player modeling is relevant to this work as AI gameplaying agents are used in place of humans for the sake of rapidly studying the efficacy of this method.

Observations of human play data has been used to bias tree search agents to play card games \cite{devlin2016combining}. Outside of learning from human data, several projects have demonstrated that human behavior can be mimicked by limiting computational resources~\cite{zook2015monte,nelson2016investigating}.
Khalifa et al.~\cite{khalifa2016modifying} identify another method that can be used to manipulate tree search to act more like humans. In this work, we utilize player modeling by using gameplaying agents in lieu of human players to demonstrate the method's efficacy.

\section{General Video Game Artificial Intelligence Framework (GVG-AI)}

GVG-AI is a research framework for general video game playing~\cite{perez2016general,perez2019general}, aimed at exploring the problem of creating artificial players capable of playing a variety of games. Organizers host an annual competition where AI agents are scored on their performance in unseen games. Each agent is given 40 milliseconds to submit an action provided with a forward model for the current game. 
The framework's environment has grown over years~\cite{perez2019general}, including new competition tracks such as level generation~\cite{khalifa2016general}, rule generation~\cite{khalifa2017general}, learning agents~\cite{torrado2018deep}, and two-player agents~\cite{gaina2016general}. To date, the framework contains a diverse set of games numbering over 100, including familiar titles such as \emph{Pacman} (Namco 1980) and \emph{Sokoban} (Imabayashi 1981), and brand new games such as \emph{Wait For Breakfast}.
% What is it, cite overview papers, competition papers

\section{Game Mechanic Alignment Theory}\label{sec:formal}
This paper presents our theory of ``Game Mechanic Alignment''.
%julian shouldn't it be Game Mechanics Alignment?
Within this framework, game mechanics can be categorized in terms of the \emph{system} (the game) and the \emph{agent} engaging with it.

Usually, games contain designer-defined reward systems which are consistent regardless of who is playing. The impact these reward systems have on play and the form they take highly depends on the conscious decisions of the designers. In a general sense, these reward systems can be interpreted as systemic penalties versus systemic rewards, which usually guide the player toward winning and away from losing. This is a separate concept from the environmental reward explicitly defined in reinforcement learning environments~\cite{sutton2018reinforcement}. In games that do not have explicit winning conditions, such as Minecraft (Mojang, 2008) or The Sims (Maxis, 2000), this condition can be substituted with another win-like condition, such as defeating the Ender Dragon in Minecraft, getting a job in The Sims, having $X$ amount of happy customers in RollerCoaster Tycoon (Chris Sawyer, 1999), etc.
% Mechanics that score positively on this axis lead to greater player empowerment, rewards, or winning. Mechanics that score negatively lead to disempowerment, penalties, and perhaps losing the game. 

In addition to built-in systemic rewards, player-specific incentives also influence how a player engages with the game. For example, speed runners and casual players have very different goals and motivations. A speed runner may bypass, skip, or glitch their way to the end of the game. A casual player may take their time and explore the environment. They may even test out losing to better understand how certain failure mechanics work. Thus, a player's goals may be independent of the game's systemic rewards. A game may reward a player when they move toward the right side of the screen, but a player may want to first collect every powerup before moving on to the next checkpoint. In an extreme case, a player interested in exploring the losing mechanics of a game may find themselves moving counter to the systemic rewards. If agential incentives and systemic rewards are in agreement for a specific mechanic, we consider this mechanic to be \emph{in alignment}.

We can think of mechanics as existing in a 2D space according to how they correlate to systemic reward systems and agential incentives. The extremes of the ``Systemic Rewards'' axis would coincide with critical and fatal mechanics as defined by previous work in automated game mechanic discovery~\cite{green2020automatic}:

\begin{itemize}
    \item \emph{Critical Mechanics}: The set of game mechanics which must be triggered to win, or the equivalent to winning.
    \item \emph{Fatal Mechanics}: The set of game mechanics which result in losing, or the equivalent harshest environmental penalty.
\end{itemize}

At the extremes of the ``Agential Incentives'' axis lie mechanics that the player feels interanally incentivized to pursue or to avoid:

\begin{itemize}
    \item \emph{Incentive Mechanics}: The set of mechanics that a specific player is incentivized to trigger over the course of the level to accomplish that player's subgoals.
    \item \emph{Avoidance Mechanics}: the opposite of incentive mechanics. The set of mechanics that a specific player avoids triggering over the course of the level to accomplish that player's subgoals.
\end{itemize}

% Drawing upon theories proposed in previous automated game mechanic discovery work~\cite{green2020automatic}, there exist key points within a 2D space of environment extrinsic rewards and player-specific intrinsic motivations:
% \begin{itemize}
%     \item \emph{Critical Mechanics}: The set of game mechanics which must be triggered to win, or the equivalent to winning.
%     \item \emph{Fatal Mechanics}: The set of game mechanics which result in losing, or the equivalent harshest environmental penalty.
%     \item \emph{Incentive Mechanics}: The set of mechanics that a specific player is incentivized to trigger over the course of the level to accomplish that player's subgoals.
%     \item \emph{Avoidance Mechanics}: the opposite of incentive mechanics. The set of mechanics that a specific player avoids triggering over the course of the level to accomplish that player's subgoals.
%     \item \emph{Neutral Mechanics}: the set of mechanics that do not fall into any of the above mechanic sets.
% \end{itemize}

\begin{figure}
    \centering
    \includegraphics[width=\columnwidth]{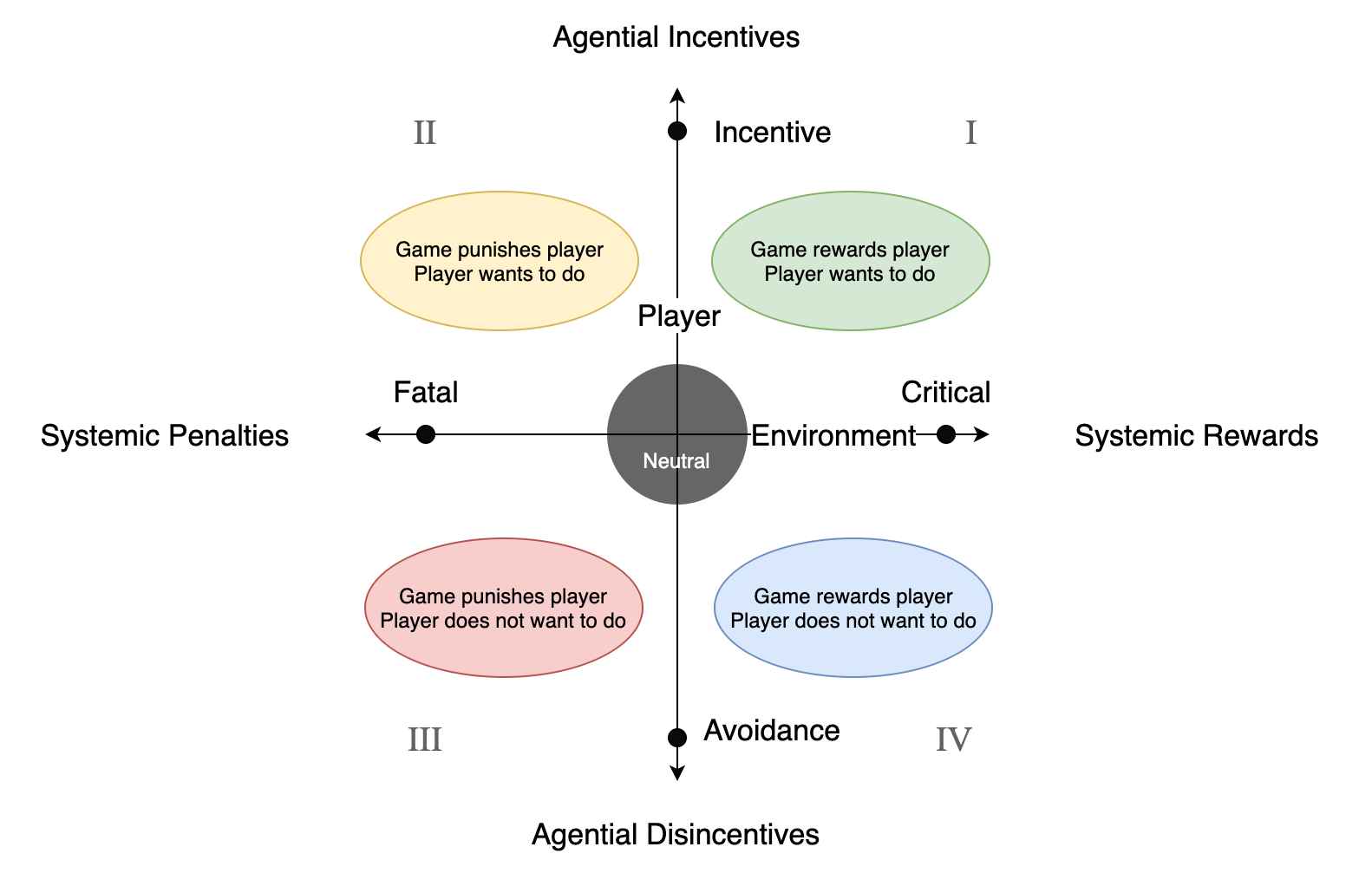}
    \caption{Game Mechanic Alignment Theory} %philip name and quadrant names
    \label{fig:mechanic_theory}
\end{figure}

The intersection of agential incentives and systemic reward axes creates an origin point, aka the ``neutral'' zone, where the player's behavior is neither influenced by systemic reward/penalty or agential motivations/aversions or only marginally influenced. The quadrant that a mechanic inhabits will embody the relationship that a mechanic has in terms of the environment and the player. Figure~\ref{fig:mechanic_theory} shows the player incentives and systemic rewards space plotted as 2D axes where x-axis represents systemic rewards and the y-axis represents agential incentives. These two axes divide the space into four quadrants. They are, in counter-clockwise order:
\begin{itemize}
    \item Quadrant 1 (Green): Both the environmental rewards and the player's motivations encourage the player to trigger this mechanic.
    \item Quadrant 2 (Yellow): The environment punishes the player but the player wants to trigger this mechanic anyways.
    \item Quadrant 3 (Red): Both the systemic rewards and the agential motivations discourage the player from triggering this mechanic.
    \item Quadrant 4 (Blue): The environment rewards the player but the player wants to avoid triggering this mechanic regardless.
\end{itemize}

When game mechanics are within Quadrant 1 or Quadrant 3 (green and red zones), they can be considered to be ``in alignment''. In other words, both the environment and the player are in agreement in regards to rewards and incentives. Perfect alignment would be at the $y=x$ line. However, if mechanics lie within Quadrant 2 and/or Quadrant 4 (yellow and blue zones), they are player-environment misaligned. The player may be motivated to take actions that cause environmental penalties, or else refuse to take actions that would otherwise give them rewards. It is important to note that being ``in alignment'' is not a judgement of player ability, nor is it a statement on what is ``correct'' behavior in the game. It is simply a way to categorize mechanics relative to the systemic rewards of the game and how often the player is motivated to trigger them during play.

In the following sections, we propose how this theory can be applied in practical situations. Section \ref{sec:examples} describes how a game designer might apply this theory to some video game examples. Section \ref{sec:methods} explains a methodology which can be used to estimate mechanic alignment using playtrace data.

\section{Video Game Examples}\label{sec:examples}
To demonstrate the usefulness of this framework, in this section we present examples using several well-known games: \emph{Super Mario Bros} (Nintendo 1985), \emph{Minecraft} (Mojang 2009), and \emph{Bioshock} (Bioware 2007). For each game, we highlight aspects relevant to the discussion of player incentives and systemic rewards and provide mechanic alignment charts as examples of how player incentives associated with certain actions could differ for two hypothetical player profiles. Since these examples are simply for illustration purposes, the values on the x and y axis are meant to be taken qualitatively. In each example, we simplify the number and description of the game mechanics: each has many more mechanics and they may be described subjectively different depending on the individual doing analysis. In sections~\ref{sec:methods} and~\ref{sec:experiments} we propose a method for estimating these values given playtraces of a game or level by many different players. 

\subsection{Super Mario Bros}

While much has been said about how the Mario series teaches the player through careful level design~\cite{mariotut2015euro}, a lot of the discussion around the series focuses on critical and fatal mechanics such as moving to the right, collecting power ups and jumping over hazards. These mechanics directly help the player progress towards winning states or avoid losing states. However, the game also acknowledges and incorporates in its design various agential motivations that players are likely to exhibit.

For example, the scoring system featured in the early games of the franchise rewards the players both for actions that directly help the player progress and avoid losing, such as killing enemies and picking up power-ups, and also for actions that display mastery over the game which have high appeal to advanced players, such as grabbing the flagpole at a higher spot and finishing the level faster. 

% \begin{figure}
%     \centering
%     \includegraphics[width=0.4\columnwidth]{images/real_examples/mario_guidance_foreshadowing.png}
%     \caption{The coins at this section of level 1-2 in Super Mario World both \textit{guide} the player's attention toward the upper portion of the screen and \textit{foreshadow} the path that will be enabled if the P switch is pressed.}
%     \label{fig:mario_world}
% \end{figure}

The collectible coins fulfill multiple roles. At first, they provide no immediate benefit other than serving as a token of player-motivated achievements such as mastering a tricky jump or uncovering secrets. But once a certain number of coins are collected, the player is given an extra life, which helps avoid a game-over state. Finally, coins and other collectibles in Mario and similar games can be used to lead the player guide the player toward secrets or suggest alternate paths that may require a power-up, which Khalifa et al.~\cite{khalifa2019level} call \emph{guidance} and \emph{foreshadowing}, respectively. Following these cues can lead to both systemic rewards and fulfilling player goals.

\begin{figure}
    \centering
    \includegraphics[width=\columnwidth]{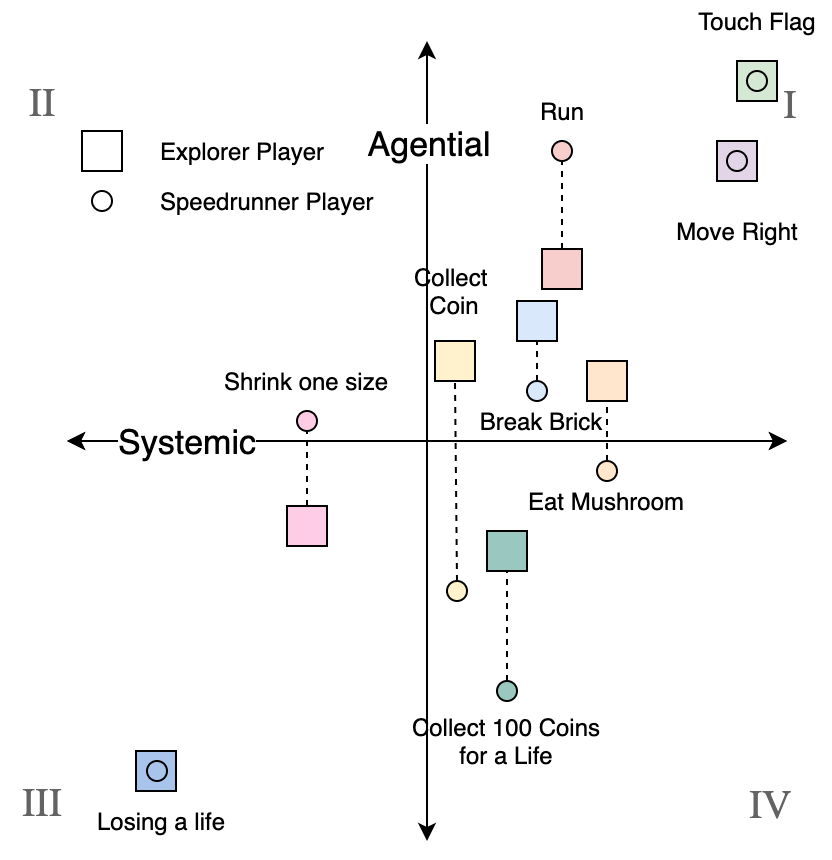}
    \caption{An example of alignment axes for two different players in Super Mario Bros}
    \label{fig:mario_chart}
\end{figure}

% Figure~\ref{fig:mario_world} shows an example of guidance and foreshadowing in \emph{Super Mario World} (Nintendo 1990).

While there is no explicit in-game tutorial for these reward systems, taken as a whole they show that designers can take steps to align systemic rewards with the actions that satisfy the player's motivations. Figure~\ref{fig:mario_chart} suggests a possible Agential-Systemic Mechanical Alignment chart for the mechanics of the game and the hypothetical player profiles of ``explorer'' and ``speedrunner''.  Both players are ultimately interested in beating the level, but the explorer player does this in a slower-paced and safer way. The explorer enjoys eating Mushrooms for extra safety, breaking Bricks out of curiosity for what's inside, and collecting coins that are easily accessible. The speedrunner, as the name suggests, wants to beat the level with the lowest possible in-game time and will avoid collectibles and bricks unless this results in a faster clear. The speedrunner will even occasionally take damage on purpose to shrink size in order to go through narrow paths, making its y-axis value higher than the explorer player. While neither player wants to purposefully waste time, the explorer will only choose to run when it is safe to do so, while the speedrunner will run most of the time, even at the risk of losing a life (and thus restarting the level and the timer). That is why the speed runner has a higher value for ``Run'' mechanic over explorer. 

For the x-axis values, we assigned relative order for these mechanics for what we think the game designer wants the player to do to finish the game. That's why ``Touch Flag'' and ``Move Right'' have the highest x-axis value, as they provide progress toward finishing the game, while ``Losing Life'' and ``Shrink one size'' have the lowest values on the x-axis as they prevent progress.

\subsection{Minecraft}

\begin{figure}
    \centering
    \includegraphics[width=\columnwidth]{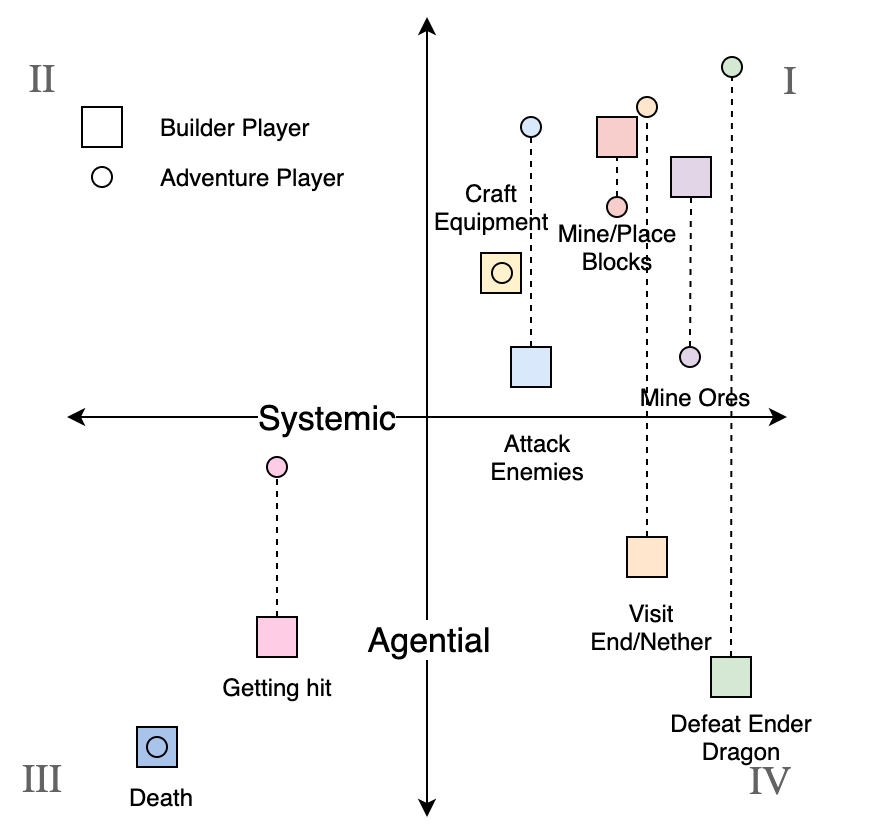}
    \caption{An example of alignment axes for two different players in Minecraft}
    \label{fig:minecraft_axes}
\end{figure}

Minecraft presents an interesting case of study for our framework since much of the appeal of the game comes from fulfilling player-incentivized goals. While triggering the credits by beating the Ender Dragon could be considered ``winning'' the game in a traditional sense, players are free to ignore this goal (potentially indefinitely) and focus on exploring, mining, crafting and building structures. 

Figure~\ref{fig:minecraft_axes} illustrates the mechanical alignment for two hypothetical player profiles: the ``builder'', who takes enjoyment from building structures, and the ``adventurer'', who seeks challenging mob encounters, including the Ender Dragon and other opponents found in the Nether and the End dimensions. Both players place equal value on crafting equipment as it enables both in reaching their goals. Neither player desires to be hit, but the adventurer is more comfortable with the possibility, and gets hit more often as consequence of the more frequent combat encounters.

For the x-axis values, we assigned them from the perspective of the designer who wants the player to reach the credit scene and finishing the game. As you can see, all mechanics provide progress toward reaching this goal are on the extreme right side such as ``Visit End/Nether,'' ``Mine Ores,'' and of course Defeating the Ender Dragon. While all the mechanics that hinders this progress lies on the extreme left such as ``Getting hit'' and ``Death.'' Minecraft contains hundreds of block types, and several different types of ores. To simplify this exercise, we have elected to merge all blocks and ores into generic mechanics. However, in theory a designer could make each of these their own mechanic for more granular analysis.

\subsection{Bioshock}

% \begin{figure}
%     \centering
%     \includegraphics[width=0.4\columnwidth]{images/real_examples/bioshock.png}
%     \caption{A tutorial in \textit{Bioshock: Remastered} explaining the consequences of a choice: choosing to harvest a Little Sister yields a higher immediate environmental reward (ADAM) but results in the death of a child. Choosing to save them yields a smaller immediate environmental reward but can be intrinsically rewarding to players who want to play in a ``morally good'' way.}
%     \label{fig:bioshock}
% \end{figure}

\begin{figure}
    \centering
    \includegraphics[width=\columnwidth]{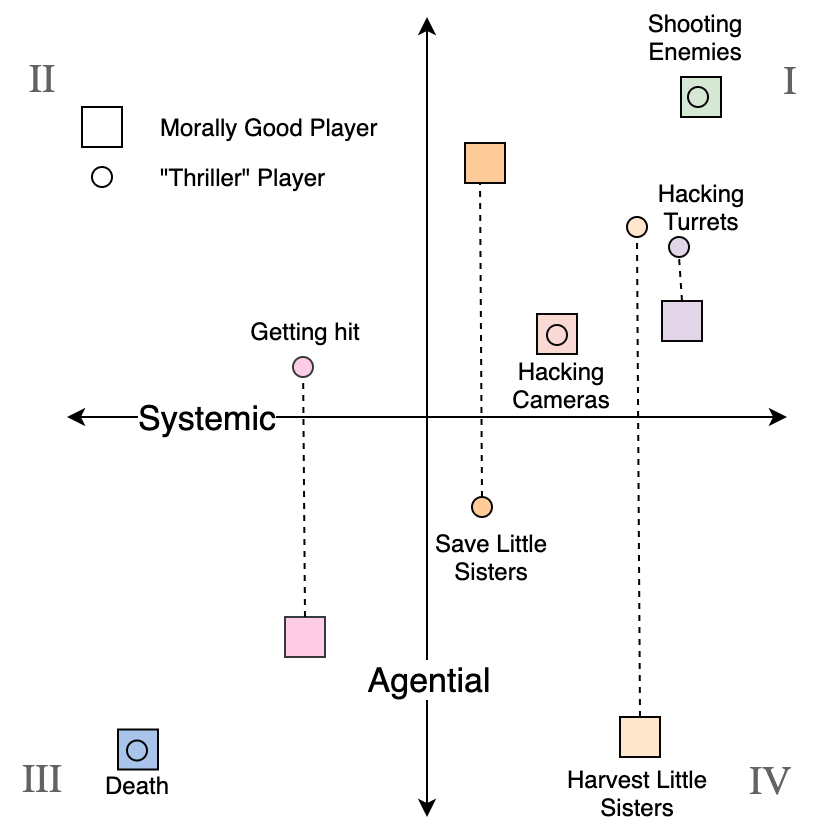}
    \caption{An example of alignment axes for two different players in Bioshock}
    \label{fig:bioshock_axes}
\end{figure}

In \textit{Bioshock}, the player is faced with an important choice regarding the fate of characters known as \textit{Little Sisters} which incurs in both moral (player incentivizes) and systemic implications. When meeting one of these genetically-modified young girls, the player can choose to either harvest or save them. Harvesting them yields more ADAM, a systemic reward that serves as in-game currency for upgrades, but results in the child's death. Saving them yields less ADAM but can create other environmental benefits and can lead to a ``happier'' ending. Thus the game pits a player's motivations for taking a moral path and reaching the ``happy'' ending against environmental considerations.

% A textual tutorial explains the immediate consequences of each option the first time the player has to choose between them~\ref{fig:bioshock}.

Figure~\ref{fig:bioshock_axes} illustrates a possible alignment chart for two hypothetical player profiles: the ``morally good'' player and the ``thriller'' player. Both these players have the same final goal which is beating the game, but the morally good player is trying to reach it with the least amount of killing, while the thriller only cares about power and destruction. That is why they differs on the ``Little Sisters'', the thriller since it needs more power, it is highly motivated on killing the sisters as they provide too much power compared to saving them. On the other hand, the morally good player don't care about the power but feels bad about killing these little girls (even if they are not real) and they might be motivated to reach the good ending of the game.

Similarly to all the previous games, we sort the mechanics on the x-axis such as mechanics near the right are pushing the player to progress in the game such as ``Shooting Enemies'' and ``Hacking Turrets''. While, mechanics near the left blocks that progress and cause the player to replay certain areas such as ``Death'' and ``Getting hit''.

% \subsection{Spider-Man}

% A big aspiration for players of \textit{Spider-Man} is to feel like the namesake hero. One of the addresses this aspiration is through the many aerial movements the player can execute while web-swinging through Manhattan. Some of these movements allow for faster traversal but others, called ``Air Tricks'' are there mostly for cosmetic purposes. However, as discussed previously, it is often in the interest of game creators to make sure that environmental and intrinsic rewards are aligned. In this case, the creators chose to not only reward these tricks with a small amount of experience (for character upgrades) and focus (to enable special moves) but also provided an explicit in-game tutorial for the mechanic~\ref{fig:spiderman}.

% (Rodrigo) Examples of tutorials for critical mechanics are plentiful. Fatal mechanics are less often directly taught, but there are some interesting examples: the behavior of a new enemy can be revealed by having them kill an NPC, or the game will sometimes require you to fail in a task before teaching you how to beat it (need to think of specific examples).

% \begin{figure}
%     \centering
%     \includegraphics[width=0.5\columnwidth]{images/real_examples/spiderman.jpeg}
%     \caption{A tutorial for aerial tricks in \textit{Marvel's Spiderman}. These tricks are largely cosmetic and offer a very small environmental reward, but can be of high intrinsic value for certain players.}
%     \label{fig:spiderman}
% \end{figure}

\section{Computational Estimation of Mechanic Alignment}\label{sec:methods}
In this section, we propose a method for automatically estimating agential incentives and systemic rewards for the game mechanics from a certain game and level. This method uses a sizable distribution of playtraces from a diverse set of artifical agents/human players where each agent/player plays the same level multiple times. The large group of player data allows for the influence of player incentives to be separated out from the environment specific motivations and thus mechanic alignments to be calculated for many types of players. Similar to any statistical technique, the bigger and more diverse the input data, the more accurate are the estimations.

Without loss of generality, the method to calculate how critical/fatal a mechanic is is fairly straightforward and intuitive. We simply want to know how often a mechanic occurs in a winning/losing playtrace as opposed to how often it occurs in any playtrace in general. For example, a critical mechanic would occur in every winning playtrace while likely occurring infrequently in losing playtraces (winning/losing doesn't need to be literal, it could be a certain condition that the player needs to trigger as discussed in section~\ref{sec:formal}). The frequency of a mechanic across playtraces can be viewed as a probability density function (PDF) that represents how likely a mechanic is to occur in a playtrace. It then follows that the difference between the PDF of the mechanic and the PDF of the mechanic given winning/losing would give a quantitative measure of how critical a mechanic is. This can be applied directly to player motivations by simply conditioning the mechanic on playtraces of the given player. We use the Wasserstein distance~\cite{wasserstein1969markov} to calculate the distance between the two distributions. The following equation shows the general form of the distance calculation.
%For both rewards, we follow the same method to estimate them. The method measures the distance between the mechanic distributions (number of times a certain mechanic happen in a playtrace) under a certain condition with respect to the full distribution of that mechanic. For the Environment reward (x-axis), the condition is winning the game (winning doesn't need to be literal winning the game, it could be a certain condition that the player need to pass as discussed in section~\ref{sec:formal}), while for the Intrinsic reward (y-axis), the condition is a certain agent(s) is being used. We decided to use the first Wasserstein distance~\cite{wasserstein1969markov} as our distance calculation due to its simplicity and not having any prior assumption about the distributions. The following equation shows the general form of the distance calculation.
\begin{equation}\label{eq:distance}
    D_{m,c} = W_{1}(PDF(m|c), PDF(m))
\end{equation}
where $m$ is the current mechanic, $c$ is the current condition ($win$ in case of systemic rewards and $agent$ in case of agential motivations), and $W_{1}$ is the first Wasserstein distance, $PDF(m|c)$ is the distribution of the mechanic $m$ given the condition $c$ happening, and $PDF(m)$ is the distribution of the mechanic $m$ in all the playtraces.

The PDF can be calculated directly from the discrete data. The frequencies for each playtrace can be normalized to form a rough, discrete, approximation of the true PDF. The Wasserstein distance can then be computed directly on this discrete data to compute the estimated distance. This approach has the benefit of not needing to fit any distribution to the data and therefore does not require any assumptions about the data distribution. Calculating the result over a PDF also would allow for this approach to be used when the PDF is known through other means. 

% Equation~\ref{eq:distance} doesn't take in consideration the scale of each mechanic. For example, some mechanics could be happening only once per playtrace such as finding a certain key, while other mechanics could be happening multiple times such as killing enemies. Without scaling the distance metrics ($D_{m,c}$), the distance metrics are not comparable as the result of Wasserstein distance is relative to the magnitude of each mechanic. To solve this problem, we scale the distance metrics ($D_{m,c}$) using the maximum number of times that mechanic $m$ can be fired ($Max_{m}$). The following equation transforms the distance metric from equation~\ref{eq:distance} to the normalized distance metric ($N_{m,c}$).
% \begin{equation}\label{eq:normalized}
%     N_{m,c} = \frac{D_{m,c}}{Max_{m}}
% \end{equation}

The distance proposed in equation~\ref{eq:distance} is a scalar with value between 0 and 1, and doesn't expression direction in the space.
To calculate the direction, we compare the mean of $PDF(m)$. If the conditional distribution has a higher average, it means it is more encouraged to happen, while if it has lower average, it is discouraged from happening.
\begin{equation}\label{eq:direction}
    S_{m,c} = Sign(\mu_{PDF(m|c)} - \mu_{PDF(m)})
\end{equation}

To calculate the final rewards, Systemic reward ($E_{m}$) and Agential reward ($I_{m}$), we combine both equations ~\ref{eq:direction} and ~\ref{eq:distance} to get the final values. Equation~\ref{eq:final} shows the final equation for both rewards where the difference is the agential incentive estimation is conditioned on a certain agent(s) are playing, while the systemic reward is condition on having a winning playtraces.
\begin{equation}\label{eq:final}
    \begin{aligned}
    I_{m} &= S_{m,agent} \cdot D_{m,agent}\\
    E_{m} &= S_{m,win} \cdot D_{m,win}
    \end{aligned}
\end{equation}

\section{Experimental Evaluation}\label{sec:experiments}

\begin{figure*}
    \centering
    \begin{subfigure}[t]{.28\linewidth}
        \centering
        \includegraphics[height=2.8cm]{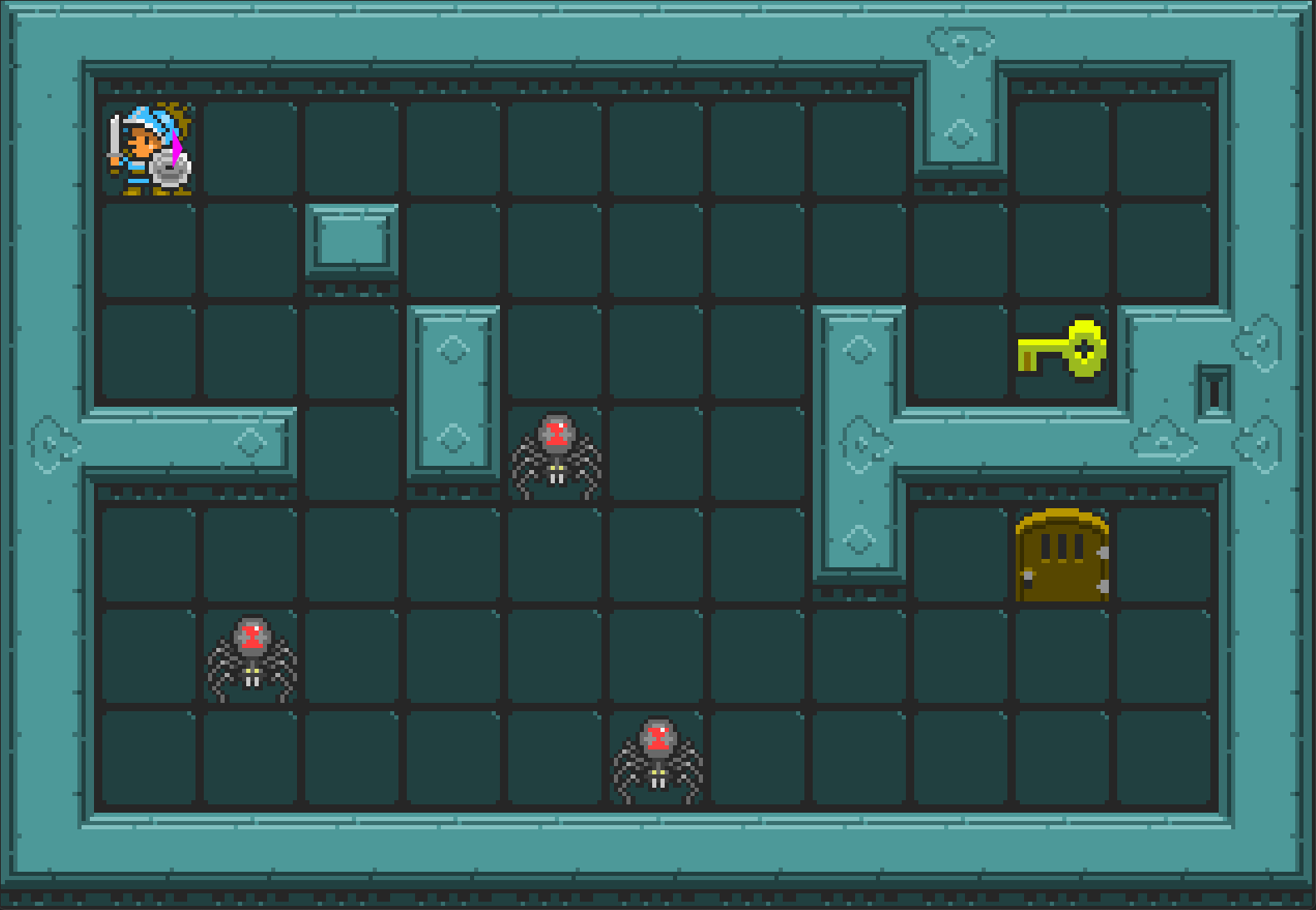}
        \caption{Zelda}
        \label{fig:zelda_level}
    \end{subfigure}
    \begin{subfigure}[t]{.48\linewidth}
        \centering
        \includegraphics[height=2.8cm]{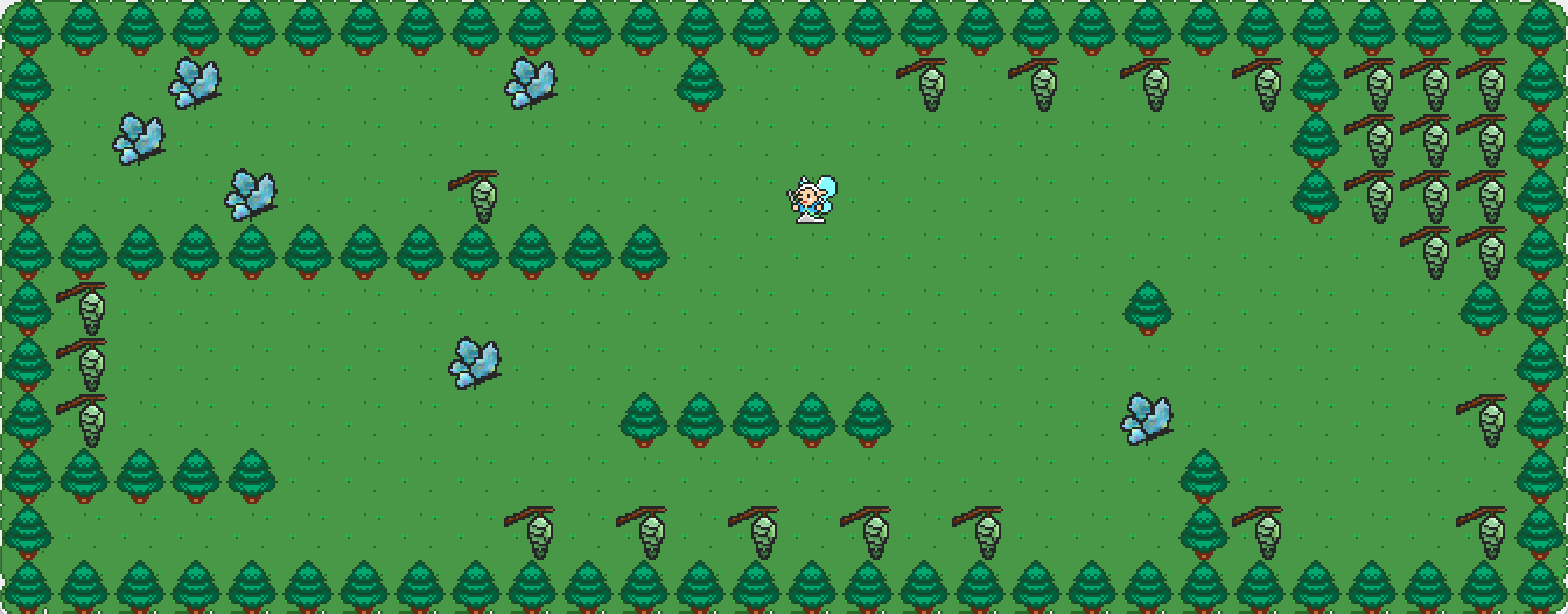}
        \caption{Butterflies}
        \label{fig:butterflies_level}
    \end{subfigure}
    \begin{subfigure}[t]{.21\linewidth}
        \centering
        \includegraphics[height=2.8cm]{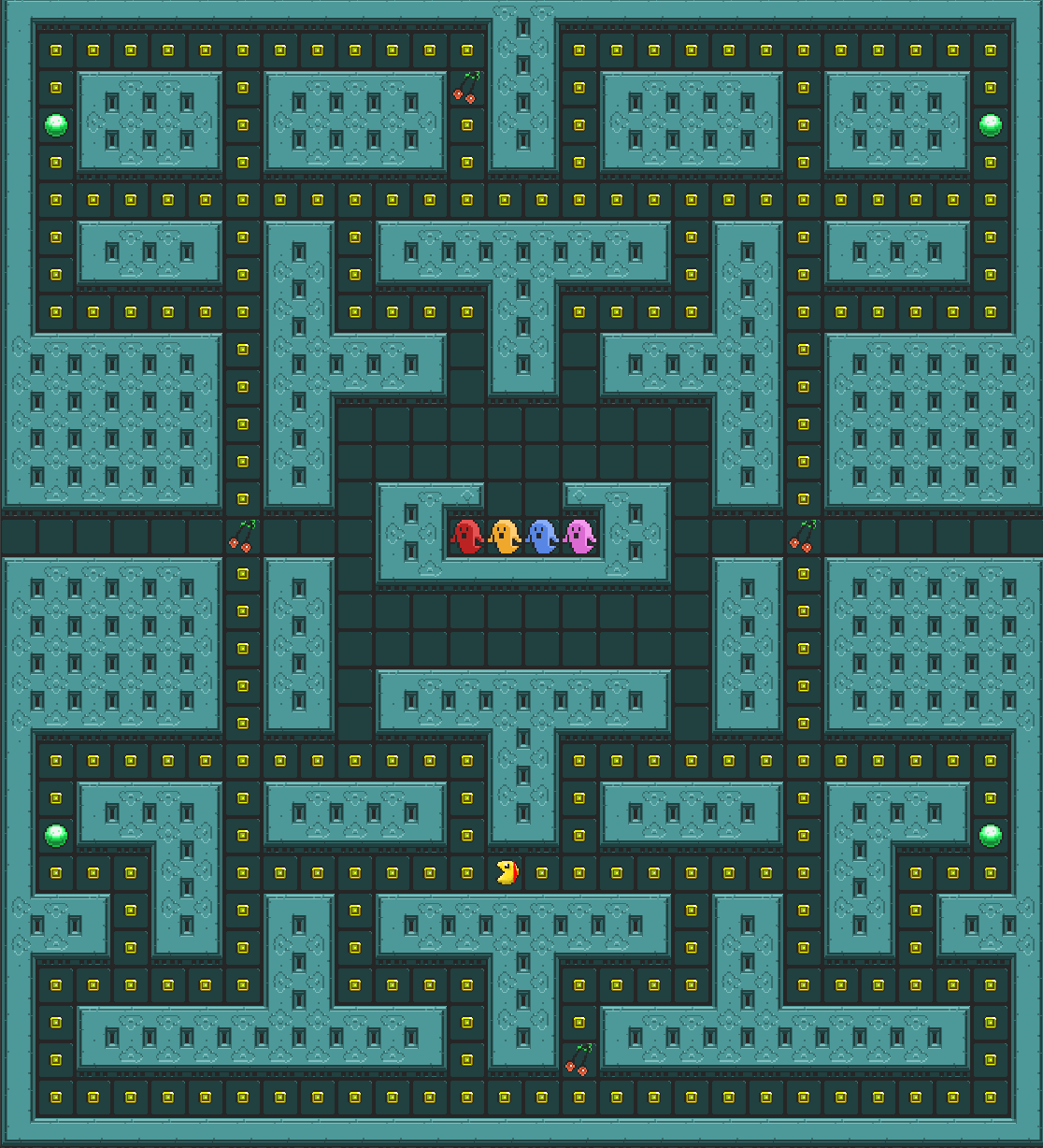}
        \caption{Pacman}
        \label{fig:pacman_level}
    \end{subfigure}
    \caption{GVGAI game levels used to test our estimation method.}
    \label{fig:gvgai_levels}
\end{figure*}

We ran a diverse set of $26$ agents on $3$ game levels from the General Video Game Artificial Intelligence Framework (GVGAI). The set of agents come with the public version of the framework.
%julian how did we select those agents?
One of these game levels is $Zelda$, a demake of The Legend of Zelda (Nintendo 1986) dungeon system. The other is $Pacman$, a port of Pacman (Namco 1980). The last one is $Butterflies$, which is a new game where the player tries to collect all the butterflies before all the cocoons hatches. Figure~\ref{fig:gvgai_levels} shows the levels used from each of these three games.
All 26 agents play each game 100 times for a total of 2600 playtraces per game level. We recorded every mechanic triggered by an agent, such as movement, collisions, and (in the case of Zelda) swinging a sword. Presented below are the results of 4 of these agents, which we believe best showcase a variety of playstyles to demonstrate the efficacy of our method: Adrienctx, Monte Carlo Tree Search (MCTS), a Greedy Search agent, and Do Nothing, which always takes the same action (neutral). These 4 agents are sorted based on their performance on these games~\cite{bontrager2016matching}. Adrienctx uses Open Loop Expectimax Tree Search (OLETS) algorithm to play the games, it is also a previous winner of the planning track in the GVGAI competition~\cite{perez20152014}. MCTS algorithm comes with the GVGAI framework where it is a vanilla implementation of Upper Confidence Bounds of Trees (UCT) algorithm~\cite{browne2012survey}. Greedy Search agent looks only for one step ahead and pick the action that will either make it win or increase the score. Finally, Do Nothing agent, as its name implies, just stands still without executing any actions until it dies or the game times out.

\subsection{Zelda}

\begin{figure}
    \centering
    \includegraphics[width=\columnwidth]{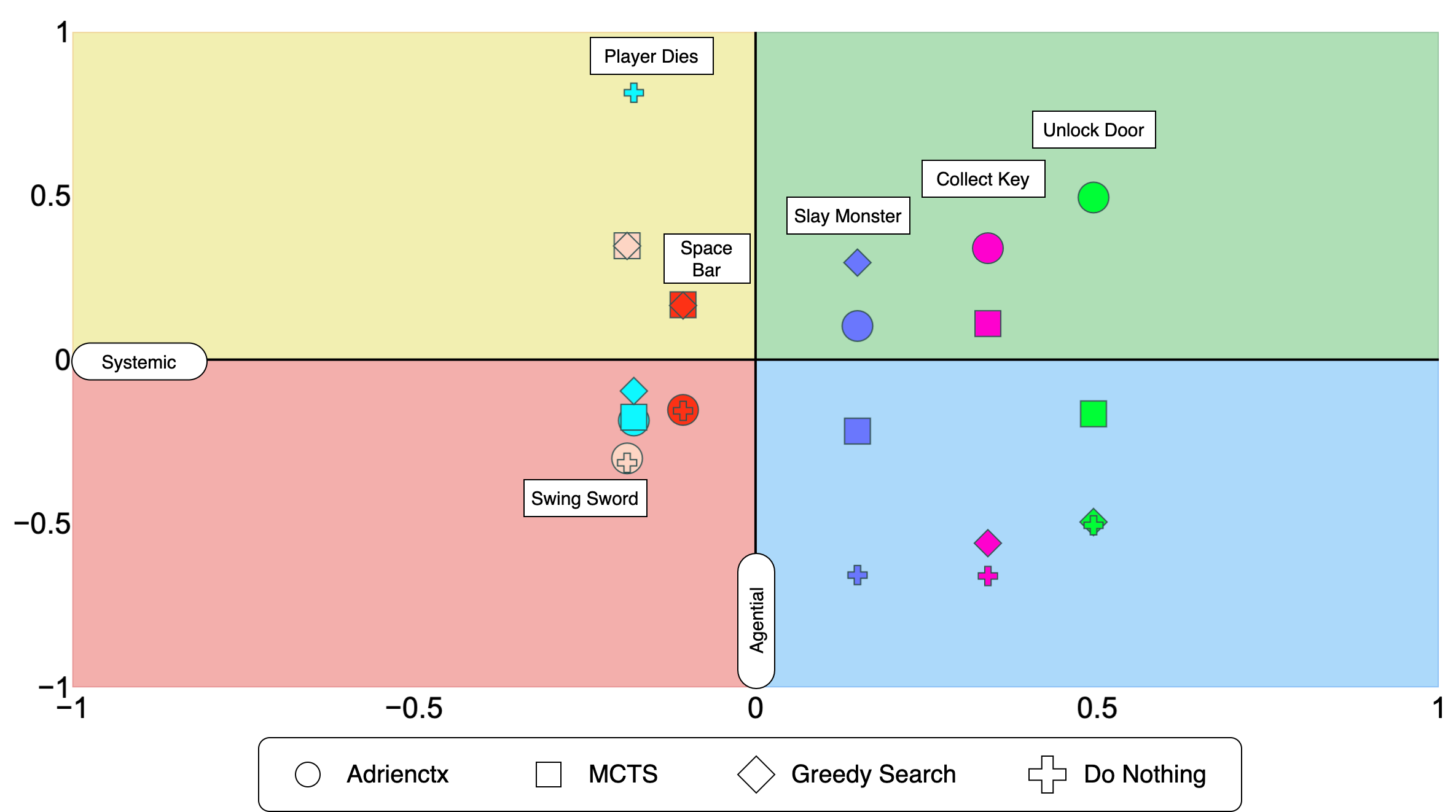}
    \caption{The Agential-Systemic Mechanical Alignment graph for $4$ agents on GVGAI's Zelda.}
    \label{fig:zelda}
\end{figure}

The goal of Zelda is to collect a key and unlock the door on the far side of the level (as seen in figure~\ref{fig:zelda_level}). Along the way the player encounters monsters, who can destroy the player if they collide. The player can swing a sword in front of them, destroying any monster it touches. Monsters move randomly around the level every couple of game ticks. Figure \ref{fig:zelda} displays the mechanical alignment of the four agents.

``Unlock Door'' and ``Collect Key'' are the two most systemically rewarding mechanics, which makes sense considering Zelda's nature as an adventure game. ``Slay Monster'' follows behind them as also positively rewarding. Interestingly enough, ``Space Bar'' (when the player presses the space bar) and ``Swing Sword'' are negatively rewarding. The discrepancy between the two mechanic scores can be explained if we consider the frequency that these mechanics are triggered. The sword can only be swung every couple of seconds due to the frame animation for the attack. Thus pressing space and swinging a sword are counted as separate mechanics, as pressing the space bar does not guarantee the sword will be swung. These mechanics are scored negatively in the environmental measurement. Agents who chase after monsters to slay them tend to also die from the stochastic nature of the game. Because monsters move randomly, it is impossible for an agent to accurately predict if it will move and in what direction at any particular game tick. Slaying monsters requires the player to risk themselves by being right next to one.

Adrienctx (circle) far outperforms other agents when it comes to collecting the key and unlocking the door. In fact, Adrienctx appears to be aligned along $y=x$, and its mechanics are in perfect agential-systemic alignment.  Greedy Search (diamond) seems more inclined to chase after and slay monsters than the other agents, while the Do Nothing (cross) agent does nothing at all. The MCTS (square) agent is the least inclined to slay monsters while being often slain by them.

We can use this graph to analyze the agent's gameplay performance like a human player:
\begin{itemize}
    \item Adrienctx understands the game mechanics very well. They could take more risks and kill more monsters to get a higher score. It could be that they enjoy the maze side of the game and would enjoy more challenging levels with complex wall patterns
    
    \item MCTS knows to collect the key but not what to do with it after. They also do not seem inclined to slay monsters, although they seem to avoid them well enough. They might need an easier level.
    
    \item Greedy Search could benefit from learning to collect the key and unlock the door, but they seem to be heavily motivated to slay monsters. They might enjoy levels that contain large quantities of monsters for a greater challenge. 
    
    \item Do Nothing is entirely out of alignment with the systemic rewards. We could assume that they are purposefully dying a lot. But most likely this is because the player may not understand the game rules. Perhaps they should be given the opportunity to play a very simple series tutorial levels that teaches them the basic mechanics.
\end{itemize}

\subsection{Butterflies}

\begin{figure}
    \centering
    \includegraphics[width=\columnwidth]{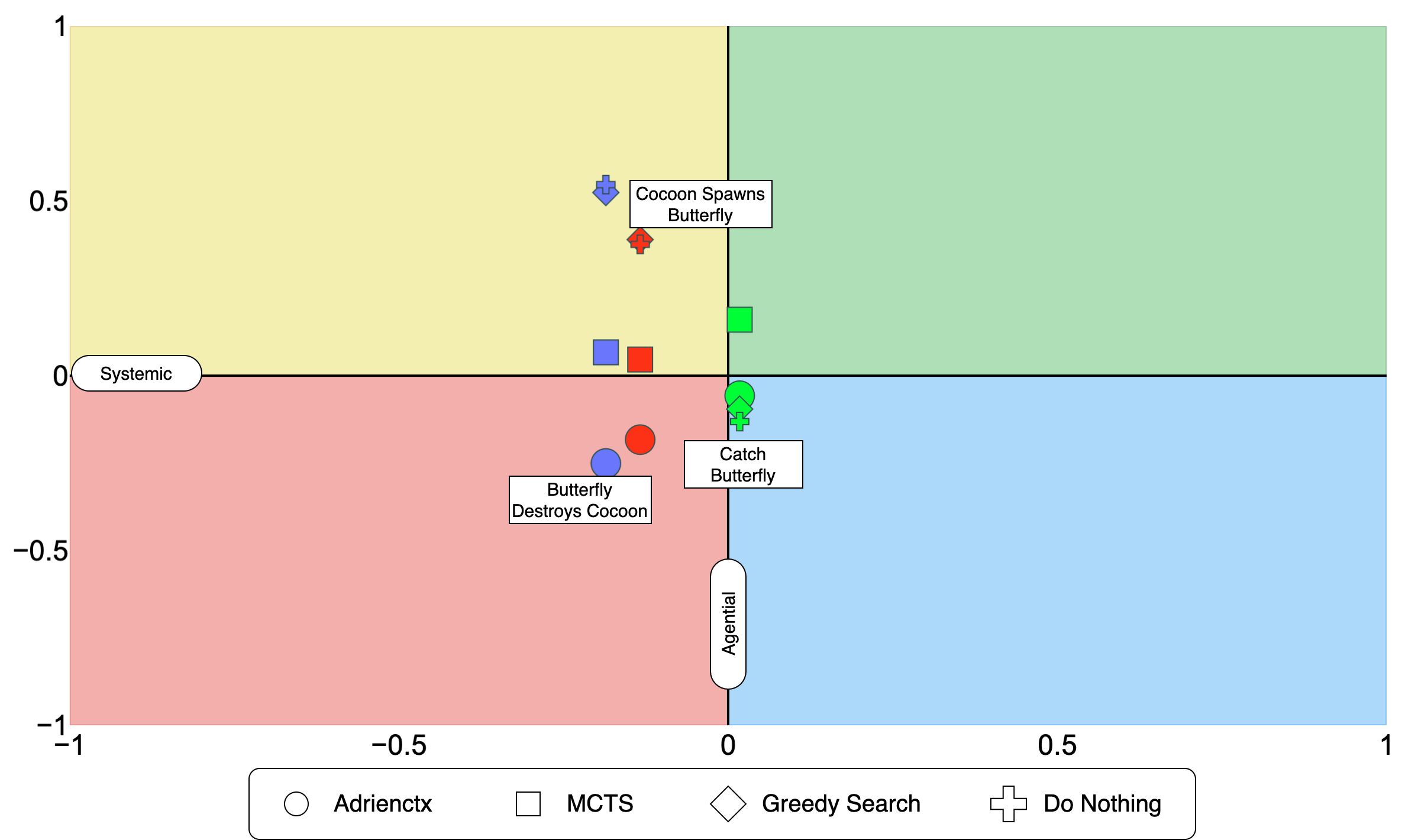}
    \caption{The Agential-Systemic Mechanical Alignment graph for $4$ agents on GVGAI's Butterflies.}
    \label{fig:butterflies}
\end{figure}

Butterflies is considered a ``deceptive game''~\cite{anderson2018deceptive}, i.e. one where the reward structure is designed to lead a player away from globally optimal decisions. The goal of butterflies is to clear the level of all butterflies, which fly around randomly (figure~\ref{fig:butterflies_level}). More butterflies can be spawned from cocoons, which crack open after a butterfly touches it. However, the player will lose if no more cocoons are in the level. Therefore, the optimal strategy is to let all but one cocoon open to spawn as many butterflies as possible and collect them all.
Figure \ref{fig:butterflies} displays the mechanical alignment of the four agents.

Due to the deceptive nature of the game, two of the mechanics (cocoons bursting and spawning butterflies) are heavily associated with losing. Remember that if all the cocoons of the game pop, the player will lose. Most agents do not to engage with the risk-reward of spawning more butterflies for a higher score, and therefore cocoons do not open nearly as often in winning playtraces. Collecting a butterfly is slightly positively associated, as it happens slightly more in winning playtraces.

Adrienctx is in near perfect alignment: while ``catch butterflies'' is just under the x axis, it tries to catch all the butterflies as soon as possible which make it catch less butterflies overall compared to the MCTS agent. MCTS collects a lot of butterflies as it allow cocoons to burst into new butterflies (that is why that ``Cocoon spawns butterflies'' and ``Butterfly Destory Cocoon'' are slightly positive on the y-axis). This behavior of MCTS is probably due to being less efficient compared to Adrienctx in collecting butterflies. Because it does not collect them as quickly, more butterflies causes more cocoons to pop, creating even more butterflies and allowing the agent to get a higher score. The rest of the agents (``Do Nothing'' and ``Greedy Search'') do not have any mechanics in alignment (all of their cocoons would eventually pop and they would lose the game).

Similar to Zelda, we could look at the graph and analyze these agents like a human player:
\begin{itemize}
    \item Adrienctx appears to not let many cocoons burst open, which limits its ability to score high. They may be trying to play it safe and not risk losing the game. They also might not understand that by waiting a little longer before catching all the butterflies so they have higher score. They might benefit from a prompt reminding them that cocoons can be good in the short term.
    \item MCTS lets lots of cocoons burst open for a greater challenge. They may be more risk taking and not afraid of losing. They may appreciate playing larger levels with lots of cocoons.
    \item Neither Greedy Search nor Do Nothing seem to understand the mechanics of the game and could benefit greatly from an easier level and a tutorial.
\end{itemize}

\subsection{Pacman}

\begin{figure}
    \centering
    \includegraphics[width=\columnwidth]{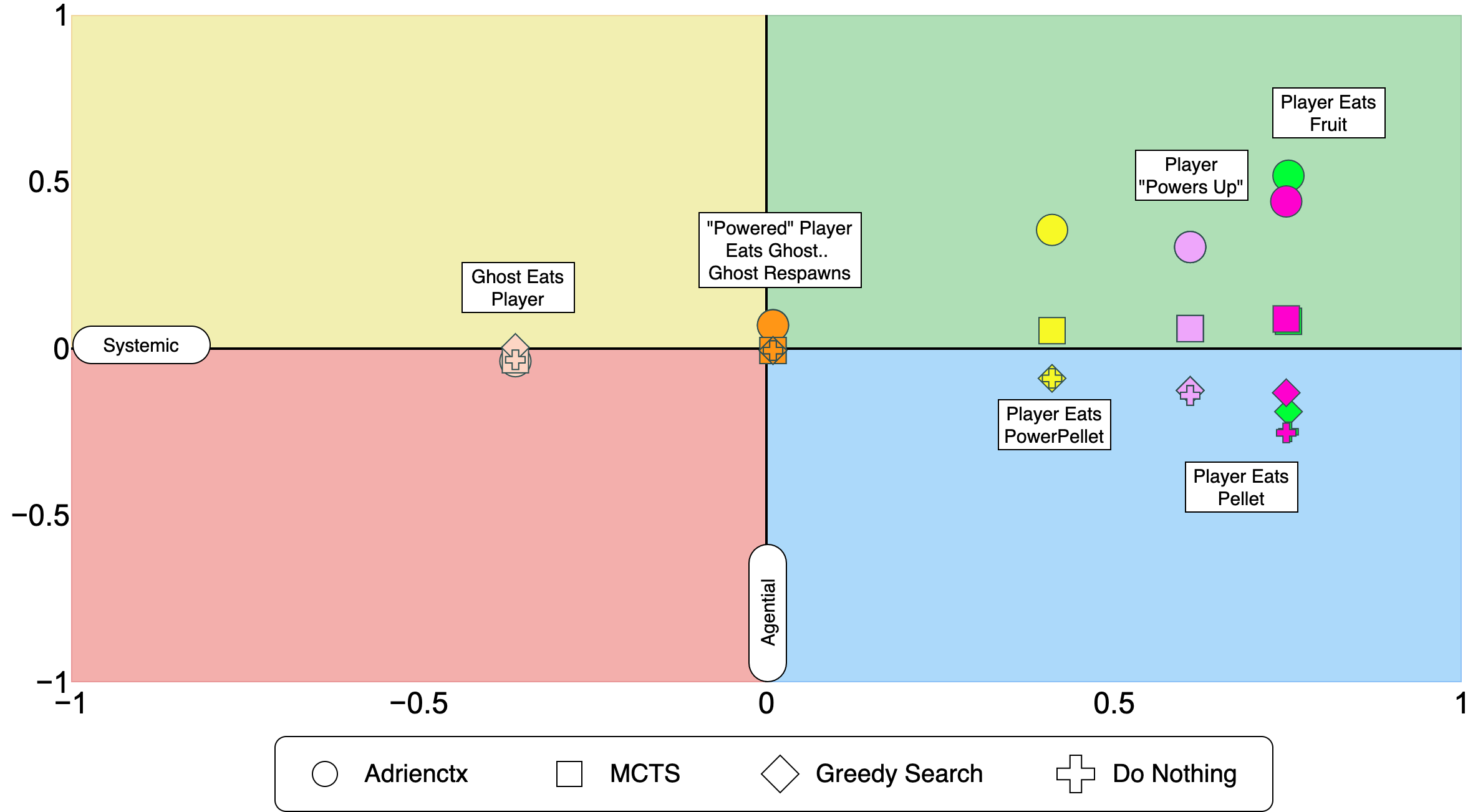}
    \caption{The Agential-Systemic Mechanical Alignment graph for $4$ agents on GVGAI's Pacman.}
    \label{fig:pacman}
\end{figure}

The Pacman level is made up of a 2-dimensional maze, which the player traverses while collecting all the pellets and fruit in the level (figure~\ref{fig:pacman_level}). Four deadly ghosts chase the player in the maze and must be avoided. Some of the pellets are ``power pellets'' which grant the player a temporary invulnerability to the ghosts, allowing the player to eat them and force them to respawn in the center of the maze. Figure \ref{fig:pacman} displays the mechanical alignment of the four agents.

Eating fruit pellets, and power-pellets are the highest systemically rewarding mechanics of the game, which is in line with its winning conditions. Eating ghosts are slightly positively rewarding, suggesting that this is not as aligned with winning and could be avoided to win. Getting eaten by a ghost results in the player losing, so it makes sense that it is heavily penalizing.

Adrienctx seems to understand the basic premise of Pacman, in that it must eat the pellets and fruit to win. However, all four agents seem nearly equivalent in getting eaten by ghosts, suggesting that none of them can reliably escape losing relative to each other. This is supported by the fact that there are only $9$ winning playtraces out of the $2600$. Adrienctx is relatively more capable of collecting power-pellets and proceeding to eat ghosts. DoNothing, on the other hand, is incapable of performing any of these actions. GreedySearch is only slightly better than DoNothing in this regard. MCTS, however, seems to be capable of engaging in these aspects of the game, however not nearly as effectively as Adrienctx.

Looking at these agents as human players, we could deduce the following:
\begin{itemize}
    \item Adrienctx's incentives are in alignment with the systemic rewards of Pacman. They seem to lose nearly as often as other players though, so they are might playing levels that match their skill level.
    \item MCTS does not seem to be nearly as successful in collecting pellets or fruits. They might need less ghosts or maybe more power pellets.
    \item Greedy Search and Do Nothing is almost entirely out of alignment. It could be that they may not understand the mechanics of the game. It could also be that they are incentivized to perform some other activity in the game, like leading the ghosts around in a specific pattern or seeing how long they can survive without power pellets.
\end{itemize}

\section{Discussion}

Game Mechanic Alignment allows us to visualise 
how different players engage with various game mechanics according to their internal incentives and systemic rewards. This has many potential applications, such as categorizing players based on the y-value of different events  (e.g. through clustering), validating a designer's assumptions of what features of the game will be most enjoyed by players, building reward systems that are aligned (to the desired extent) with player's preferences and building tutorials that support actions preferred by different types of players, even if they are not all equally rewarded by the reward systems.

It is important, however, to distinguish between the theoretical notion that agential incentives might be more or less aligned with systemic rewards and the quantitative methods used to estimate this alignment. The first, and most obvious limitation of the quantitative estimation done in this paper is that we use artificial agents as proxies for what a player's behavior might look like. More accurate estimations could be done with either human play traces or using artificial agents that attempt to emulate a specific persona. However, these agents are suitable proxies for this experiment, as they are intentionally built to utilize different incentive systems during play (Adrienctx uses Hierarchical Open-Loop Optimistic Planning, MCTS uses a systemic-reward biased tree, Greedy has no search and takes the best immediate action, etc). By using agents which were not emulating a specific persona, we have demonstrated with noisier data that the method can distinguish well between various players.

Another limitation of the estimation is that it relies on the correlation between triggering a mechanic and winning the game (for the x axis) or between triggering a mechanic and belonging to a certain player profile (for the y axis). These correlations are not necessarily causal/intentional and could be affected by spurious factors. One such factor is skill. Consider a hypothetical level where a cosmetic item with no functional value is hidden near the  exit. The event of picking up this item would have a high correlation with winning, and therefore would show up far to the right in the x-axis, even though it does not improve a player's chance to beat the level. And if two player profiles A and B are equally internally incentivized to pick up the item, but profile A reaches the end of the level more often, the event would show up higher on the y-axis for profile A, as profile A simply had more opportunities to trigger it due to its higher player skill. This can be seen in our examples, where all the automated agents except do-nothing are designed with the goal to win the level but have different skill levels, causing differences in the y-axis even with no discernible differences in motivation. 

The \emph{Talin}~\cite{aytemiz2018talin} system arguably touches upon this more directly. In Talin, mechanics are dynamically taught to the player based on their skill level. Mechanic mastery is represented using scalar values initialized by the game designer. As the player plays and either triggers or fails to trigger that mechanic, its value rises or falls. A similar system could be used to measure player skill/mastery of particular mechanics to explain how much of a y-value difference between different players is a result of their skill versus incentives. If we can be sure of the method's accuracy, it may be more beneficial to designate another axis to account for player skill. However, we want to make it clear that being unable to differentiate between incentive or skill when it comes to player behavior does not negate the utility of this technique. Two experiences could be offered to the player to counter this: one being a tutorial giving an easier level to help ``practice'', the other giving the player a more challenging level with more complex uses of the mechanic to provide entertainment. Regardless of it being due to skill or incentives, the game system/designer has a deeper understanding of what the player was doing during play than without the technique. 

Idiosyncrasies of a particular game or level could also contribute to spurious correlations. For example, if another hypothetical level contains a bifurcation where the player can choose between a red path heading North and a blue path heading South, and these paths feature opportunities to trigger different sets of events, then a naive analysis of the prevalence of these events might be confounded by an aesthetic preference for the red or blue color (or, as is more likely for bots, by an arbitrary tie-breaker between the North and South directions).

Level difficulty also may play a role here. In our examples, all agents were compared on the same level. Therefore, we can be assured that difficulty remains constant. However, many games present the player with procedurally generated sets of levels, such as Spelunky (Derek Yu, 2008). In games like these, a designer may not be able to make the assumption that all levels are equally difficult. If we assume there exists a method for measuring level difficulty, then we could test our agents only against the levels with similar difficulty and the same set of game mechanics.

% Another limitation is our y-axis estimation depends on the collected data. The estimation only cares about the collected data and not about the actual motivation of the player/agent. For example, in our experiments all the used agents have the same heuristic (win the level) which makes all the agents have the same intrinsic motivation but because each agent uses different algorithm to play the levels, they end not performing the same as each algorithm is better in playing certain games than others. This is not the problem of using agents, this can still happen with actual human players. What the player intended to do and what they reflect in the data might not be the same. When we think about all the different reason that could contribute to that problem, we figured out three main ones:
% \begin{itemize}
%     \item \textbf{Player Skill doesn't match Game Difficulty:} if the player have the intention to win the game but the game is too difficult for them, the data will show that they are always trying to lose in every playtrace which will make our method estimate that this player is motivated to lose.
%     \item \textbf{Level design}: sometimes the designers could hide certain mechanics behind others. For example: all treasures in the game are hidden after an enemy fight. If the player intention is only to collect treasures, our method will also detect killing enemies highly on y-axis.
%     \item 
% \end{itemize}

It is possible, in principle, to attempt to correct for spurious correlations. For example, if we know an event can only be triggered at a certain point of the map, we could consider only play traces that got within a certain distance of that point. Or we could, at each time-step of the play trace, simulate the game for a number of steps with the goal of triggering the event, and consider only play traces where it was possible to do so. We could also perform A/B tests where some features of the level (e.g. the events on each path) are kept constant where others (e.g. the colors) are swapped. But these corrections would come at the cost of domain knowledge and/or computational power, which is why they were not considered in our experiments.

All of our examples in Sections \ref{sec:formal} and \ref{sec:experiments} refer to games with mechanics that have either immediate or delayed rewards/penalties. Let us consider the game Fallout 2 (Bethesda 1998), a first person shooter set in a post-apocalyptic Oregon, United States. In the game, the player may encounter a consumable substance called ``Jet,'' a highly addictive meta-amphetamine which grants the player temporary short-term bonuses to their combat abilities. After the initial bonus period, however, the player will suffer heavy penalties to these same skills. Additionally, the player may become addicted to the substance, requiring them to keep dosing themselves every day or suffer additional penalties. In this example, consuming Jet has opposing systemic reward/penalties, depending on the time horizon: rewarding in the short-term, penalizing in the long-term. To visualize this using Game Mechanic Alignment, we may need to introduce a third axis to represent ``time.''

One of our core motivations for this work was to find a new alternative method to critical mechanic discovery~\cite{green2020automatic} for automated tutorial generation systems that considers the player and the environment. While this method does not discover ``critical mechanics'' as defined by Green et al, it does estimate the tendencies of the player to gravitate toward particular mechanic usage/triggers (or lack there-of). The concept of critical mechanics also neglects to take different playstyles into account, as the definition allows for only a single set of critical mechanics to exist for a given level (only those needed to win). Consider the following use case: by building a variety of artificial agents with different utility functions (and therefore explicit agential incentives), a designer can quickly obtain a large collection of playtraces. Each agent would represent a potential ``playstyle.'' Then, any single human user's playtrace can be compared against the entirety of agent traces and the designer would know which archetype category(ies) best fit the player. This player could then be served a experience personalized for that playstyle. This experience can be automatically built using the information gleaned by Game Mechanic Alignment about that playstyle by creating or removing opportunities for triggering certain sets of mechanics.

The mechanic-graph restriction of the previously mentioned critical mechanic discovery methods~\cite{green2018atdelfi,green2020automatic} is not a requirement for the Game Mechanic Alignment method. Therefore, users might find it more generalized and easier to use. In our experimental section, we used the output logs from agent gameplay in the GVGAI framework. However, mechanics could be automatically detected in similar manner to detecting highlights in soccer matches~\cite{ekin2003automatic}, or by using a method of automated mechanic discovery~\cite{cook2013mechanic}.

\section{Conclusion}
In this work, we present Game Mechanic Alignment theory, a framework which enables designers to organize game mechanics in terms of the environment and a player engaging with it. We provide several well-known games as examples for how a designer may apply this theory. To demonstrate its practicality, we propose a methodology to estimate reward values for Game Mechanic Alignment, as well as an experimental evaluation using $3$ games from the GVGAI framework. We then point out shortcomings in regards to this methodology while discussing several ways they may be overcome.

By taking both player and environment into account, tutorial generators can create highly personalized tutorials or experiences. Rather than one tutorial for a single game or level, each player or playstyle could receive their unique tutorial. Players who are highly skilled can be introduced to more complex mechanics whereas novice players can be given explanations of basic controls. Furthermore, players with certain playstyles can be introduced to completely different styles, they might not have considered. Proposed future works are to validate this theory with real human players and automatically design tutorials using this methodology to organize input, first for gameplaying agents using the same methodology from Section \ref{sec:experiments} and then for human players in a formal user study. 

% Mike: I want to say something here similar to our idea with the bayesian method where we identify events instead of mechanics. We could build a classifier that uses the alignment system to try to find as much distance as possible between different agents.

\begin{acks}
Michael Cerny Green would like to thank the OriGen.AI education program for their financial support.
Rodrigo Canaan gratefully acknowledges the financial support from Honda Research Institute Europe (HRI-EU).
\end{acks}
\bibliographystyle{ACM-Reference-Format}
\bibliography{biblio}

\end{document}